# Mini-ResEmoteNet: Leveraging Knowledge Distillation for Human-Centered Design

Amna Murtada
*Dept. of Information Technology*
*University of Khartoum*
Khartoum, Sudan
amnamurtada99@gmail.com

Omnia Abdelrhman
*Dept. of Information Technology*
*University of Khartoum*
Khartoum, Sudan
omniaabdelrhman77@gmail.com

Dr. Tahani Abdalla Attia
*Dept. of Electrical and Electronic Engineering*
*University of Khartoum*
Khartoum, Sudan
tahani@uofk.edu

*Abstract*—Facial Emotion Recognition has emerged as increasingly pivotal in the domain of User Experience, notably within modern usability testing, as it facilitates a deeper comprehension of user satisfaction and engagement. This study aims to extend the ResEmoteNet model by employing a knowledge distillation framework to develop Mini-ResEmoteNet models—lightweight student models—tailored for usability testing. Experiments were conducted on the FER2013 and RAF-DB datasets to assess the efficacy of three student model architectures: Student Model A, Student Model B, and Student Model C, their development involves reducing the number of feature channels in each layer of the teacher model by approximately 50%, 75%, and 87.5% respectively. Besides demonstrating exceptional performance on the FER2013 dataset, Student Model A (E1) achieved a test accuracy of 76.33%, marking a 0.21% absolute improvement over EmoNeXt, Moreover, the results exhibit absolute improvements in terms of inference speed and memory usage during inference compared to the ResEmoteNet model, the findings indicate that the proposed methods surpass other state-of-the-art approaches.

## I. Introduction

Facial expression recognition (FER) has become a crucial field of study in computer vision with important applications in human-computer interaction, emotion analysis, and usability testing. Subtle variations in facial components such as lips, teeth, skin, hair, cheekbones, nose, face shape, eyebrows, eyes, jawline, and mouth complicate the FER task.

FER systems operate through three main stages: face acquisition, feature extraction, and emotion classification. These stages enable the detection and interpretation of human facial expressions across seven predefined emotional categories. In the face acquisition stage, methods like Faster R-CNN and YOLO [1] are employed for efficient face detection achieving high accuracy by localizing facial regions in images. In feature extraction, CNN-based models like ResNet capture hierarchical patterns from facial data, while SE networks [2] recalibrate channel importance to focus on emotionally relevant information. Vision transformers [3] offer the ability to capture long-range dependencies across facial features. In the emotion classification stage, extracted features are assigned to one of the seven emotion classes. This process is typically performed by specialized deep learning models, with multimodal approaches integrating visual and contextual data for improved accuracy [4].

## II. Related Work

FER has advanced significantly due to deep learning, but high computational costs remain challenging for real-time deployment. Computational costs refer to resources like memory usage, inference time, and parameter count. Large models like Vision Transformers and SE-enhanced ResNets require substantial resources which limits scalability in constrained environments, TABLE I. illustrates how different FER models compare in size and computational complexity.

TABLE I.   MODEL SIZE AND PARAMETERS COMPARISON OF RESEMOTENET WITH EXISTING STATE-OF-THE-ART METHODS

| FER Models | Model Size (MB) | Total Parameters |
|---|---|---|
| POSTER++ [5] | 233.27 | 58,316,594 |
| QCS [6] | 331.50 | 82,874,119 |
| FMAE [7] | 1217.31 | 304,326,632 |
| Segmentation VGG-19 [8] | 676.88 | 169,220,807 |
| EmoNeXt [9] | 122.26 | 30,564,331 |
| Ensemble ResMaskingNet [10] | 85.15 | 21,288,263 |
| **ResEmoteNet [11]** | **320.95** | **80,238,599** |

POSTER++ [5] and QCS [6] balance model size and performance, but their parameter counts (58.3M and 82.8M, respectively) still indicate significant computational costs, particularly for real-time systems. FMAE [7] represents a highly complex model, with a size exceeding 1.2 GB, while its performance is likely superior. Segmentation VGG-19 [8]is another large model, with a size of 676.88 MB. While slightly more efficient than FMAE, it remains computationally prohibitive for many applications. EmoNeXt [9] and Ensemble ResMaskingNet [10] represent lightweight models that prioritize efficiency. Ensemble ResMaskingNet, in particular, demonstrates an optimal design for low-resource environments, with only 21.3 million parameters and a size of 85.15 MB.

ResEmoteNet [11]sits between these extremes, providing a balance between high performance and computational efficiency. With a size of 320.95 MB and 80.2 million parameters, it is more resource-efficient than models like FMAE and Segmentation VGG-19 but not as lightweight as Ensemble ResMaskingNet. ResEmoteNet achieves this balance by integrating advanced features such as squeeze-and-excitation (SE) blocks and residual connections, which enhance its classification performance but add to its computational requirements as illustrated in Fig.1.

The increasing demand for real-time FER systems has prompted the development of lightweight architectures that reduce computational complexity while maintaining accuracy. MobileNet [11] significantly reduced the number of parameters and floating-point operations by implementing depthwise separable convolutions. EfficientNet [13] advanced lightweight architectures through the use of a compound scaling strategy that methodically modified network dimensions to strike a balance between accuracy and efficiency. Despite significant advancements, lightweight architectures frequently struggle to handle imbalanced datasets and subtle emotional expressions. These challenges highlight the importance of hybrid approaches that combine lightweight design principles with optimization techniques such as knowledge distillation.

Knowledge distillation, introduced by [14], reduces the computational demands of deep learning models by transferring knowledge from a large, high-performing teacher model to a smaller student model. This process enables the student model to mimic the teacher's behavior while operating with significantly reduced resources, making it particularly valuable for real-time FER applications. Loss functions, such as distillation loss and cross-entropy loss, balance imitating the teacher, and aligning with ground truth labels. T and Alpha hyper-parameter fine-tune the distillation process, ensuring accuracy and efficiency in the student model. Knowledge distillation has optimized resource-intensive models like vision transformers and SE-enhanced CNNs for real-time use [15]. It enables smaller models to inherit the strengths of their larger counterparts while improving generalizability to diverse real-world conditions, such as varying lighting and occlusions.

To close these gaps and enhance ResEmoteNet for usability testing applications, In this paper, ResEmoteNet is used as a teacher model and applies knowledge distillation to create lightweight adaptations that will improve computational performance in student models. ResEmoteNet has been chosen as it outperforms state-of-the-art models across four FER databases[11].

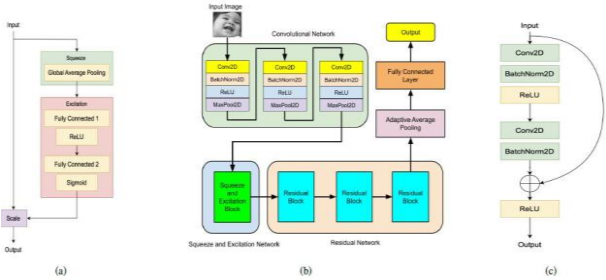

Fig. 1. Showcasing The Components of ResEmoteNet [11]

### III. METHODOLOGY

#### A. Student Models Architecture:

In the construction of the student models, we opted to reduce the parameters of the teacher model by varying proportions to examine the feasibility of parameter reduction without compromising model accuracy. This investigation led to the development of three distinct student model architectures: Student Model A involves reducing the number of feature channels (filters) in each layer of the teacher model by approximately 50%; Student Model B reduces the feature channels by approximately 75%; and Student Model C further reduces them by approximately 87.5%, as illustrated in TABLE II.

TABLE II. THE TEACHER-STUDENT MODELS ARCHITECTURE COMPARISON

| Models | Models Architecture | | | Total Parameters |
|---|---|---|---|---|
| | CNN | SE | RB | |
| ResEmoteNet | 64,128,256 | 256 | 256,512,1024 | 80,238,599 |
| Student Model (A) | 32, 64, 128 | 128 | 128, 256, 512 | 20,069,383 |
| Student Model (B) | 16,32, 64 | 64 | 64,128, 256 | 5,022,215 |
| Student Model (C) | 8,16,32 | 32 | 32,64,128 | 1,259,911 |

#### B. Our Knowledge Distillation Method:

Employing ResEmoteNet as a teacher model within the framework of knowledge distillation, the methodology prioritizes the extraction of hard labels and soft predictions from the teacher model, culminating in a composite of distillation loss functions. The input data is subsequently propagated through the student model for processing. Fig.2 illustrates the proposed method.

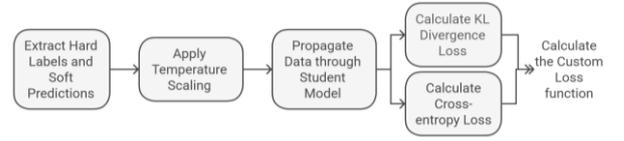

Fig. 2. The Proposed Knowledge Distillation Method

In the mechanism of Soft Target Distillation, temperature scaling is implemented via the temperature hyper-parameter T to smooth the probability distributions, thereby enhancing the efficacy of knowledge distillation. This smoothing is achieved by dividing the logits by T before the application of the softmax function. The teacher model's softened logits convey comprehensive information regarding class relationships, thereby simplifying the learning process to a certain degree. The use of the KL divergence loss facilitates the transfer of knowledge between the teacher and student soft predictions, seemingly enabling accelerated training.

$$KLDivLoss = \frac{1}{N} \sum_{i=1}^{N} \sum_{c=1}^{C} P_{I,C} . [log_{(Pi,c)} - log_{(Qi,c)}] \ (4)$$

Where $P_{i,c}$ is the "softened" probability distribution of the teacher model and $Q_{i,c}$ is the "softened" probability distribution of the student model. In the context of Hard Label Learning, the Cross-entropy loss is employed to guarantee that the student model sustains its accuracy concerning the primary task while preserving direct supervision through ground truth labels; this loss can be calculated between the predictions of the student model and the ground truth labels.

$$CrossEntropyLoss = -\frac{1}{N} \sum_{i=1}^{N} \sum_{c=1}^{C} y_{i,c} . log_{(y`i,c)} \ (5)$$

Where C is the number of classes, $y_{i,c}$ is the ground truth label for class c (1 for the correct class, 0 otherwise), $y`_{i,c}$ is the predicted probability (softmax output) for class c for sample i. Then the total loss function is a weighted combination of two components:

$$Loss = CrossEntropyLoss + \beta \times T^2 \times LossKD \ (6)$$

Where α is the weight for hard label loss, β is the weight for distillation loss, T is the temperature parameter, CrossEntropyLoss is Cross-entropy loss with hard labels and KLDivLoss is KL divergence loss between softened predictions.

## C. Datasets:

This subsection presents an overview of the datasets utilized in this study. Our experimental investigations were conducted using two datasets: FER2013 [16] and RAF-DB [17]. Our facial emotion recognition task aims to identify seven fundamental emotions: Angry, Disgust, Fear, Happy, Neutral, Sad, and Surprise. TABLE III provides a class-wise breakdown of the train-test distribution for each dataset to facilitate a comprehensive analysis

TABLE III. CLASS-WISE DATA DISTRIBUTION ACROSS: FER2013 AND RAF-DB

| Classes Distribution | FER2013 | | | RAF-DB | |
|---|---|---|---|---|---|
| | Training Set | Public Test | Private Test | Training Set | Test Set |
| Happy | 7215 | 653 | 594 | 4772 | 1185 |
| Surprise | 3171 | 56 | 55 | 1982 | 478 |
| Sad | 4830 | 895 | 879 | 2524 | 680 |
| Angry | 3995 | 467 | 491 | 705 | 162 |
| Disgust | 436 | 415 | 416 | 717 | 160 |
| Fear | 4097 | 607 | 626 | 281 | 74 |
| Neutral | 4965 | 496 | 528 | 1290 | 329 |
| **Total per set** | **28709** | **3589** | **3589** | **12271** | **3068** |

The FER2013 dataset is Ideal for benchmarking and developing models for controlled conditions. Its lower resolution and noise make it a good dataset for testing noise robustness. It was created for a facial expression recognition challenge at the ICML Workshop in Representation Learning and contains grayscale images of size 48 × 48. It is divided into training, public test, and private test sets, with annotations for the seven basic emotions.

The Real-world Affective Faces Database (RAF-DB) is advantageous for training models designed for applications in real-world contexts due to its extensive diversity, enhanced resolution, and incorporation of compound emotions. It comprises RGB facial images annotated with basic or compound expressions by a cohort of 40 independent taggers, as illustrated in Fig.3

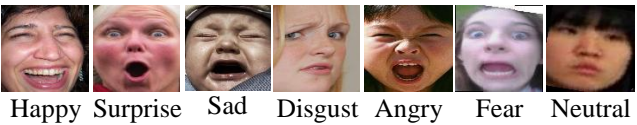

Happy  Surprise  Sad  Disgust  Angry  Fear  Neutral

Fig. 3. Showcasing Images of The 7 Emotion Classes

## D. Experimental Details

In our study, an identical experimental framework was employed when utilizing both the FER2013 and RAF-DB datasets. The training regimen for the student models integrated hyper-parameters meticulously ascertained through rigorous sensitivity analyses. The experimental arrangement comprised training batches consisting of 16 samples, with the training regimen extending over 80 epochs to align with the original teacher model's training configuration. A custom loss function was implemented for optimization purposes while Stochastic Gradient Descent (SGD) served as the primary optimization algorithm. To address the class imbalance observed in Table II across all datasets, a class-weight-biased method was employed, which assigned increased importance to underrepresented classes within the loss function computation. The data preprocessing pipeline included the incorporation of Random Horizontal Flip augmentation techniques to enhance model generalization. The training procedure commenced with an initial learning rate of $1 \times 10^{-3}$, along with an adaptive scheduler which reduced the rate by a factor of 0.1 upon stagnation in learning.

Through evaluation of temperature values 1, 2, 3, 4, and 5, it was determined that a temperature of 3 consistently resulted in superior performance. This is likely attributable to the fact that a temperature of 3 offered an ideal balance between capturing intricate class relationships and preserving a robust training signal. Furthermore, the softer target probabilities achieved at T=3 may have enabled the student model to generalize more effectively by assimilating broader class representations, thereby avoiding overfitting the teacher's most confident predictions. Additionally, experimentation with alpha values ranging from 0.10, 0.15, and 0.20 demonstrated that alphas of 0.15 and 0.20 were optimal, as the model's performance reached saturation at these points, indicating an ideal equilibrium between learning from the teacher's soft labels and the ground truth labels. This equilibrium allowed the student models to adeptly capture nuanced inter-class relationships while maintaining adherence to the hard-label classification task.

Our experimental evaluations were executed utilizing the PyTorch framework on an NVIDIA Tesla P100 GPU infrastructure provided by Kaggle. In assessing the performance of our facial emotion recognition system, we employed Accuracy (%) as the principal metric, defined as:

$$\text{Accuracy (\%)} = \frac{TP+TN}{TP+TN+FP+FN} \times 100 \quad (7) \text{ [11]}$$

where TP is True Positive, TN is True Negative, FP is False Positive, and FN is False Negative.

## IV. RESULTS AND DISCUSSION

This section delineates the outcomes of our experimental investigations conducted on the benchmark datasets FER2013 and RAF-DB. We appraised the efficacy of our proposed approach, the ResEmoteNet student models, with the testing results systematically organized in Table III for FER2013 and Table IV for RAF-DB.

### A. ResEmoteNet Student Models Performance Across Datasets:

We commenced our study with the FER2013 dataset, primarily due to its challenging characteristics arising from issues such as inaccurate labeling, the absence of faces in certain images, and an imbalanced data distribution. TABLE IV. presents the experimental results of FER2013, the initial experimentation involved student model A, configured with a temperature of 3 and an alpha value of 0.20. This model appeared to exhibit the optimal performance with a test accuracy of 76.33%, which is a reduction of 3.46% compared to the teacher model.

The next best performance was observed during the fourth experiment with student model B, which was also configured with a temperature of 3 but with an alpha of 0.15, resulting in a test accuracy of 70.20%, indicating a decrease of 9.59% relative to the teacher model. Further experimentation in the sixth trial explored the potential of student model C, comprising merely 1,259,911 parameters, which achieved an accuracy of 58.58%, thereby exhibiting a decline of 21.21% from the teacher model. Additionally, we provide the confusion matrices of the ResEmoteNet student model for the

FER2013 dataset, which illustrate the model's class-wise Confusion across the respective test sets in Fig.4.

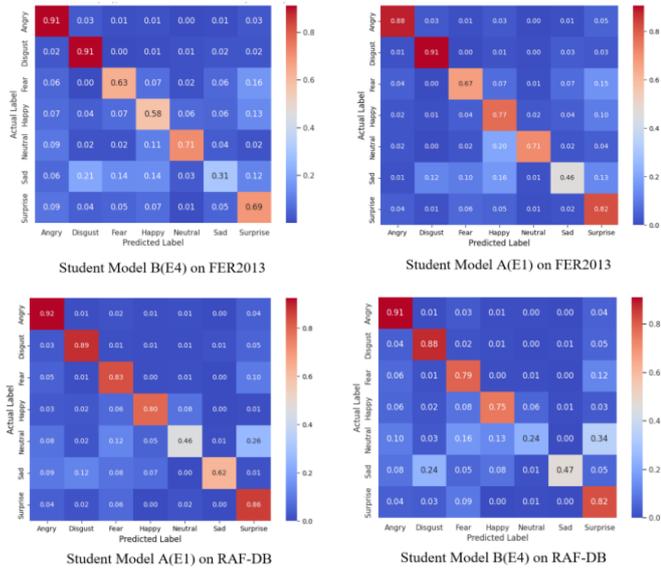

Fig. 4. The Confusion Matrices of The ResEmoteNet Student Models

TABLE IV. depicted the experimental results on RAF-DB, which is known for presenting real-world challenges including variations in pose, lighting, and occlusion, was deemed suitable for evaluating the extensive capabilities of student models A and B. Under identical hyper-parameter settings, student model A (E1) achieved a test accuracy of 84.21%, which is 10.55% lower than that of the teacher model. In comparison, student model B (E4) attained a test accuracy of 81.41%, showing a reduction of 13.35% relative to the teacher model.

TABLE IV. THE EXPERIMENTAL RESULTS CONDUCTED ON FER2013 AND RAF-DB

| Table Head | Accuracy(%) | | Accuracy (%) | Loss |
|---|---|---|---|---|
| | T | Alfa | | |
| FER2013 | | | | |
| Teacher Model (ResEmoteNet) | - | - | 79.79 | 0.056 |
| **E1: Student Model A** | **3** | **0.20** | **76.42** | 0.7590 |
| E2: Student Model A | 3 | 0.15 | 76.17 | 0.0755 |
| E3: Student Model B | 3 | 0.20 | 68.75 | 0.0970 |
| **E4: Student Model B** | **3** | **0.15** | **70.20** | 0.9659 |
| E5: Student Model B | 3 | 0.10 | 69.56 | 0.0934 |
| E6: Student Model C | 3 | 0.20 | 58.58 | 0.1202 |
| RAF-DB | | | | |
| Teacher Model (ResEmoteNet) | - | - | 94.76 | 0.1802 |
| **E1: Student Model A** | **3** | **0.20** | **85.00** | 0.5213 |
| **E4: Student Model B** | **3** | **0.15** | **82.45** | 0.6225 |

*B. Student ResEmoteNet Models Performance Comparison with Teacher ResEmoteNet Performance:*

TABLE V. presents a comparative analysis of the performance of the proposed student models against the established ResEmoteNet teacher model. The findings illustrate that student models A and B surpass the ResEmoteNet model concerning model size, memory usage during inference, and inference speed. These student models have been engineered to be lightweight alternatives to the teacher model. Specifically, the ResEmoteNet model exhibits a considerable size of 306.09 MB, while student model A is significantly smaller with a size of 76.56 MB, representing an enhancement of +74.99%. Furthermore, student model B demonstrates exceptional improvement with a size of only 19.16 MB, reflecting a reduction of +93.74%. Consequently, these student models are more adapted to real-time applications. Student model A (E1) is particularly appropriate for scenarios requiring a balance between lightness and accuracy, whereas student model B (E4) is optimal for contexts prioritizing reduced memory usage.

TABLE V. COMPARISON OF THE RESEMOTENET STUDENT MODELS WITH EXISTING RESEMOTENET TEACHER MODEL

| Characteristics | Characteristics | | |
|---|---|---|---|
| | *ResEmoteNet* | *Student Model (E1A)* | *Student Model (E4B)* |
| Model Size (MB) | 306.09 | 76.56 | 19.16 |
| *FER2013* | | | |
| Accuracy (%) | 79.79 | 76.42 | 70.20 |
| Memory Usage (MB) | 10102.94 | 5088.46 | 1814.97 |
| Average Inference Time (ms) | 1.4 | 0.14 | 0.15 |
| *RAF-DB* | | | |
| Accuracy (%) | 94.76 | 85.00 | 82.45 |
| Memory Usage (MB) | 838.19 | 827.96 | 748.72 |
| Average Inference Time (ms) | 3.2 | 0.29 | 0.24 |

In addition to demonstrating exceptional performance on the FER2013 dataset, the student model A (E1) achieved a test accuracy of 76.42% while utilizing 5088.46 MB of memory during inference. In contrast, the student model B (E4) obtained a test accuracy of 70.20% with a memory usage of 1814.97 MB during inference. These models exhibited absolute improvements in memory usage during inference by 49.63% and 82.05%, respectively, compared to the ResEmoteNet model, which consumes 10102.94 MB of memory. Furthermore, regarding inference speed, student models A (E1) and B (E2) recorded average inference times of 0.14 ms and 0.15 ms, respectively, representing absolute improvements of 90.00% and 89.29% over the ResEmoteNet teacher model, which has an inference time of 1.4 ms.

In the context of working with RAF-DB and evaluating student model performance on the Fer2013 dataset, student model A (E1) achieved a test accuracy of 76.42% with a memory usage of 827.96 MB during inference. In contrast, the student model B (E4) recorded a test accuracy of 70.20% while employing 748.72 MB of memory during inference. The student models A (E1) and B (E4) demonstrated absolute improvements over the ResEmoteNet model in terms of memory usage during inference, showing reductions of 1.22% and 10.68% respectively, compared to ResEmoteNet's 838.19 MB. Furthermore, regarding inference speed, student models A (E1) and B (E2) exhibited average inference times of 0.29 ms and 0.24 ms, respectively, representing absolute improvements of 90.94% and 92.50% over the ResEmoteNet teacher model, which has an inference time of 3.2 ms.

*C. Student ResEmoteNet Models Performance Comparison with Prior Studies:*

In TABLE VI. the efficacy of the proposed methods is evaluated against various state-of-the-art techniques, with the results indicating superior performance of our methodologies compared to existing approaches. The FER2013 dataset presents notable challenges due to issues such as inaccurate labeling, the presence of images devoid of facial features, and an imbalanced distribution of data. Despite these challenges, the student models achieved a classification accuracy of 76.33% with 20,069,383 parameters, representing a 0.21% absolute improvement over EmoNeXt [9], which operates with 30,564,331 parameters. In the context of RAF-DB, the ResEmoteNet student model A(E1) achieved a classification accuracy of 85.00%, marking a 0.2% enhancement over CMT VGGFACE [18]. Additionally, student model B(E4) attained an accuracy of 82.45%, signifying a 1.05% absolute improvement over C MT PSR [18].

TABLE VI. TEST ACCURACY (%) AND PARAMETERS COMPARISON OF RESEMOTENET STUDENT MODELS WITH EXISTING STATE-OF-THE-ART METHODS ACROSS TWO DATASETS: FER2013 AND RAF-DB.

| Table Head | Accuracy(%) | | Parameters |
|---|---|---|---|
| | *FER2013* | *RAF-DB* | |
| C MT PSR [18] | - | 81.4 [a] | - |
| C MT VGGFACE [18] | - | 84.8 [a] | - |
| POSTER++ [5] | - | 92.21 | 58,316,594 |
| QCS[6] | - | 93.02 | 82,874,119 |
| FMAE [7] | - | 93.09 | 304,326,632 |
| SegmentationVGG-19 [8] | 75.97 | - | 169,220,807 |
| EmoNeXt[9] | 76.12 | - | 30,564,331 |
| Ensemble ResMaskingNet [10] | 76.82 | - | 21,288,263 |
| ResEmoteNet[11] | 79.79 | 94.76 | 80,238,599 |
| **Student Model A(E1)** | **76.33** | **85.00** | **20,069,383** |
| **Student Model B(E4)** | **70.20** | **82.45** | **5,022,215** |

[a.] Average accuracy

## V. CONCLUSION

This study emphasizes how crucial transfer learning is to improving neural networks' ability to recognize facial emotions. We demonstrated that the student models Mini-ResEmoteNet greatly increase the overall resilience of neural networks by using the knowledge distillation approach. Our method's effectiveness is confirmed by the notable gains seen on the RAFDB and FER2013 benchmarks. Notably, Student Model A (E1) outperformed EmoNeXt by 0.21%, achieving a test accuracy of 76.33%. Additionally, Student Models A (E1) and B (E4) showed average inference durations of 0.14 and 0.15 ms, respectively, which are absolute gains over the instructor model's 1.4 ms operating time. In addition, the results demonstrate that the memory consumption during inference is increased by 49.63% and 82.05%, respectively, in comparison to the ResEmoteNet model, which uses 10,102.94 MB of memory. According to the results, the suggested techniques outperform other cutting-edge techniques. These outcomes demonstrate the efficiency of our method in precisely identifying facial emotions, providing noteworthy progress in the area of facial emotion detection. Additional improvements to Mini-ResEmoteNet models and integration of the Mini-ResEmoteNet to develop a usability testing framework will be investigated in future research.